\documentclass[10pt,twocolumn,letterpaper]{article}

\usepackage{3dv}
\usepackage{times}
\usepackage{epsfig}
\usepackage{graphicx}
\usepackage{amsmath}
\usepackage{amssymb}

\usepackage[pagebackref=true,breaklinks=true,letterpaper=true,colorlinks,bookmarks=false]{hyperref}

\usepackage{booktabs}
\usepackage{multirow}

\usepackage{mathtools}
\usepackage{siunitx}

\DeclarePairedDelimiter\abs{\lvert}{\rvert}

\DeclarePairedDelimiter\ceil{\lceil}{\rceil}
\DeclarePairedDelimiter\floor{\lfloor}{\rfloor}

\newcommand{\pluseq}{\mathrel{+}=}

\newcommand{\dbar}{d\hspace*{-0.08em}\bar{}\hspace*{0.1em}}
\newcommand*{\dv}[1]{\dbar #1}

\threedvfinalcopy

\ifthreedvfinal\fi
\begin{document}

\title{BP-MVSNet: Belief-Propagation-Layers for Multi-View-Stereo}

\author{Christian Sormann\textsuperscript{1} \\
{\tt\small christian.sormann@icg.tugraz.at}
\and
Patrick Knöbelreiter\textsuperscript{1}\\
{\tt\small knoebelreiter@icg.tugraz.at}
\and
Andreas Kuhn\textsuperscript{2} \\
{\tt\small andreas.kuhn@sony.com}
\and Mattia Rossi\textsuperscript{2}  \\
{\tt\small mattia.rossi@sony.com}
\and Thomas Pock\textsuperscript{1} \\
{\tt\small pock@icg.tugraz.at}
\and Friedrich Fraundorfer\textsuperscript{1} \\
{\tt\small fraundorfer@icg.tugraz.at} 
\and 
\textsuperscript{1}Graz University of Technology, \textsuperscript{2}Sony Europe B.V.
}
\maketitle
\thispagestyle{empty}

\begin{abstract}
In this work, we propose BP-MVSNet, a convolutional neural network (CNN)-based Multi-View-Stereo (MVS) method that uses a differentiable Conditional Random Field (CRF) layer for regularization. To this end, we propose to extend the BP layer~\cite{bp_reloaded} and add what is necessary to successfully use it in the MVS setting. We therefore show how we can calculate a normalization based on the expected 3D error, which we can then use to normalize the label jumps in the CRF. This is required to make the BP layer invariant to different scales in the MVS setting. In order to also enable fractional label jumps, we propose a differentiable interpolation step, which we embed into the computation of the pairwise term. These extensions allow us to integrate the BP layer into a multi-scale MVS network, where we continuously improve a rough initial estimate until we get high quality depth maps as a result. We evaluate the proposed BP-MVSNet in an ablation study and conduct extensive experiments on the DTU, Tanks and Temples and ETH3D data sets. The experiments show that we can significantly outperform the baseline and achieve state-of-the-art results.
\end{abstract} 

\section{Introduction} \label{sec_intro}
The goal of a dense 3D reconstruction system is to initially estimate dense depth maps for a set of overlapping input images capturing an arbitrary scene by utilizing MVS. These dense depth maps are then fused into a dense point cloud. Thus, for each given input image, we want to compute a depth estimate for every pixel. Additional inputs of an MVS pipeline are the camera poses of each image and the camera calibration, which are obtained using a Structure-from-Motion (SfM) pipeline. Following MVSNet~\cite{mvsnet}, we choose a reference view and a given set of source images having a substantial overlap with the reference image in terms of the viewed geometry. Analogous to the two view stereo case, we need to search along the epipolar lines in all source images using a similarity measure to find the correct depth value. 

\begin{figure}[t]
\centering
\includegraphics[width=0.45\columnwidth]{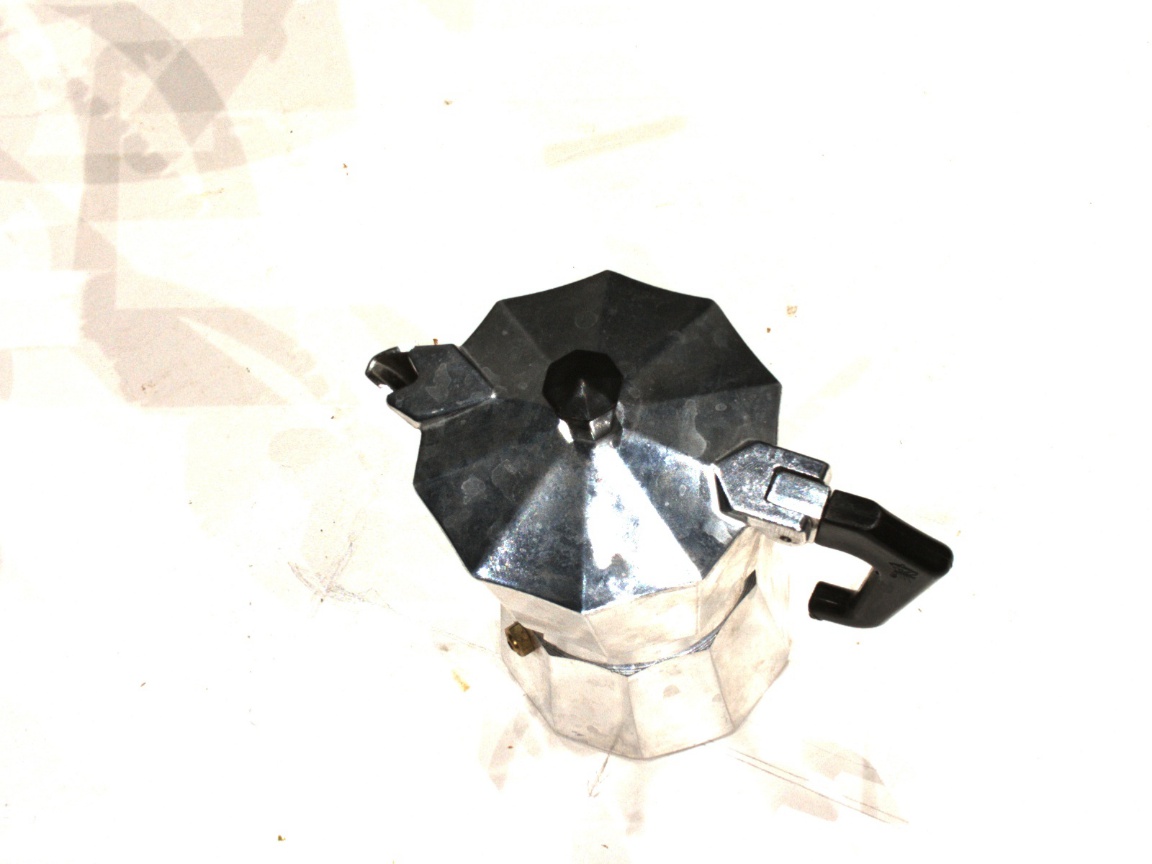}
\includegraphics[width=0.45\columnwidth]{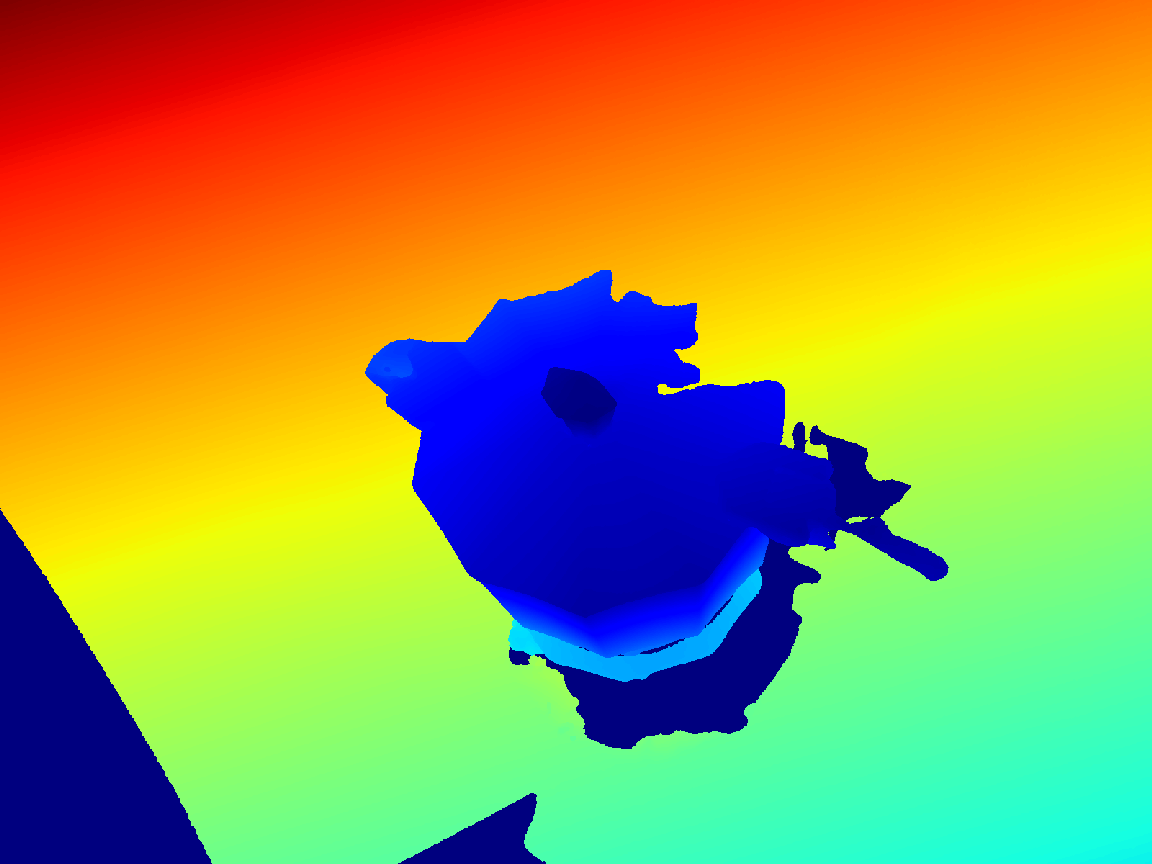} 
\includegraphics[width=0.45\columnwidth]{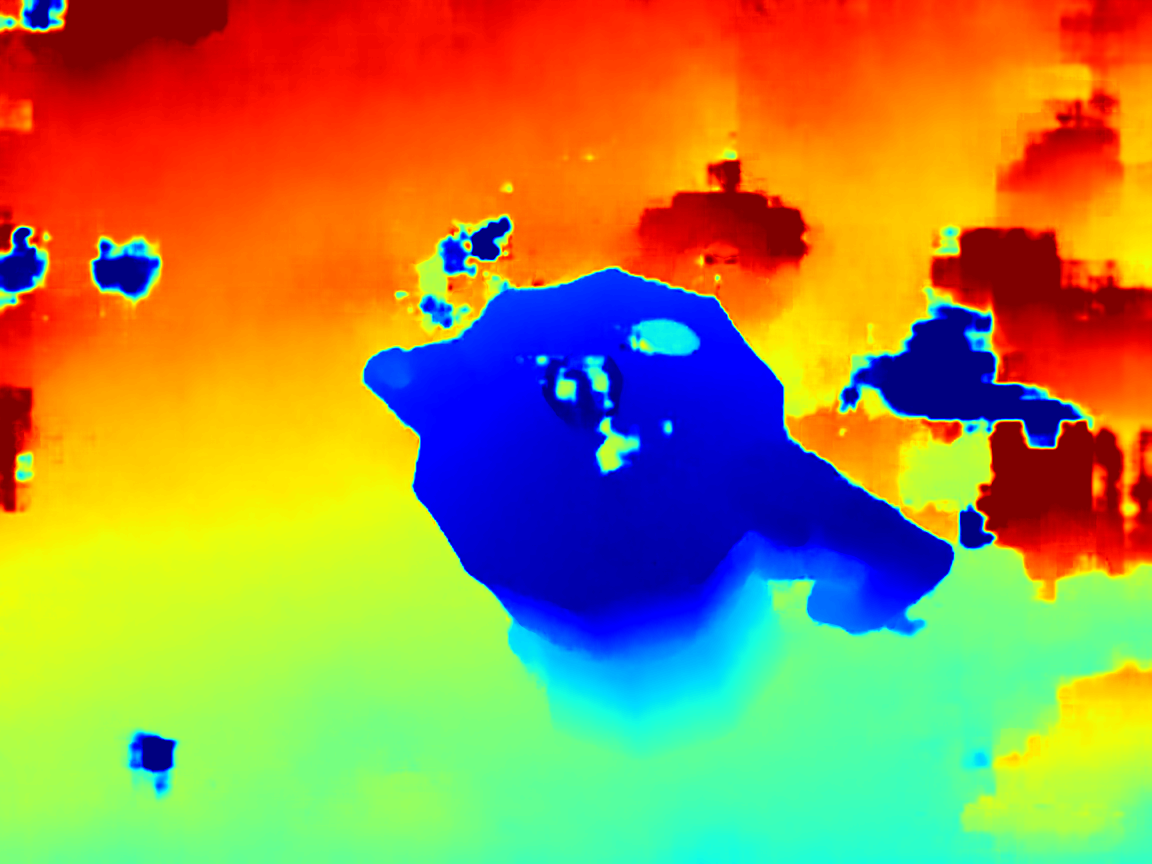}
\includegraphics[width=0.45\columnwidth]{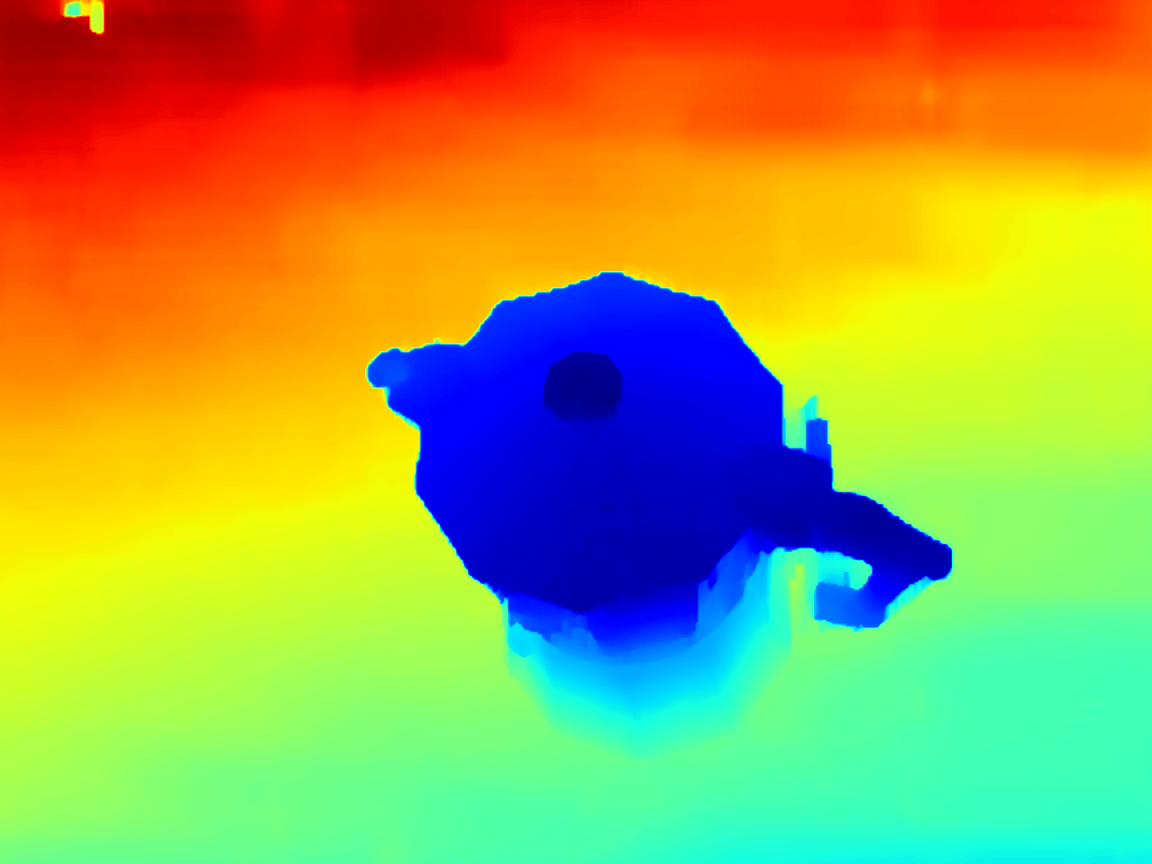}
\caption{Top left: input rgb image. Top right: ground truth~\cite{mvsnet}. Bottom left: result from CasMVSNet~\cite{casmvs}. Bottom right: result from BP-MVSNet (ours).}
\label{fig_teaser}
\end{figure}
Traditional MVS methods usually use raw color information or hand-crafted features to describe local image patches. These features are then used with a traditional similarity measure such as normalized cross correlation (NCC) to find corresponding points. However, color information alone is often ambiguous and therefore a suboptimal choice for comparing patches. Therefore \cite {mvsnet} proposed to learn optimal features for matching using a CNN and also to replace the traditional similarity measure with a learned one.
However, the authors of~\cite {mvsnet} do not use any smoothness assumption on the resulting depth maps, which is commonly used in the canonical two-view stereo~\cite{Scharstein07learnin,knoebelreiter_cvpr2017}. \cite{mvscrf} observed this and they were the first to suggest that the depth maps should also be regularized with a CRF in the MVS setting. However, the CRF inference in~\cite{mvscrf} has not been adapted for the MVS setting by explicitly handling different scales. Further, the employed mean-field approximation is highly dependent on the number of iterations. To overcome these limitations, we propose to extend the differentiable BP layer of~\cite {bp_reloaded} to enable its applicability in the MVS setting. The BP layer is a fully differentiable CRF inference layer that can deliver high quality results after a single iteration. We summarize our core contributions in the upcoming paragraph.
\paragraph{Contribution}
We propose the end-to-end learnable BP-MVSNet, which explicitly exploits prior knowledge of the MVS task via a CRF. To this end, we use the BP layer~\cite{bp_reloaded} and extend it to meet the requirements of a MVS method. In particular, we propose three extensions that correspond to our core contributions: We (i) propose a scale-independent normalization method to enable the handling of label jumps on different scene scales, (ii) add support for fractional label jumps in the CRF using a differentiable interpolation step that is fully integrable into the learning and (iii) propose a method to automatically calculate the sampling interval of the plane hypothesis beyond the initial stage. Our modifications described in Sections~\ref{sec_quant_label_jumps} and \ref{sec_pairwise_interpol} required us to extend the forward path of~\cite{bp_reloaded}. Consequently, we also provide the required gradients to learn the parameters of the employed pairwise score function in the backward path. Thus, we are able to seamlessly integrate the BP layer into a learnable MVS network. We can significantly outperform the baseline  CasMVSNet~\cite{casmvs} and achieve state-of-the-art results on the DTU~\cite{dtu} and Tanks and Temples~\cite{tanksandtemples} benchmarks.
\section{Related Work}
We group the related work for MVS into traditional approaches and CNN based solutions. 
\subsection{Traditional MVS}
Traditional MVS systems typically employ a photometric similarity measure such as bilateral weighted NCC~\cite{colmap_mvs} for evaluating different depth hypothesis. A possible technique for generating the hypothesis is the plane-sweeping MVS~\cite{plane_sweep_mvs} approach, where the depth hypothesis for each pixel is computed by sampling a number of candidate planes in the scene. Another way of generating new depth hypothesis is via the PatchMatch~\cite{patchmatch_original, patchmatch_stereo} algorithm, where a sampling scheme is used to propagate depth hypothesis across the image. 
This algorithm has also been adapted and extended for the MVS case~\cite{pm_mvs_view_select, colmap_mvs}. 

The work of~\cite{gipuma} utilizes checkerboard-based propagation to further reduce runtime and is combined with a coarse to fine scheme by~\cite{acmm}.  Furthermore,~\cite{colmap_mvs, acmm} use a forward-backward reprojection error as an additional error term for the PatchMatch estimation. MARMVS~\cite{marmvs} additionally estimate the optimal patch scale to reduce matching ambiguities. While these methods generally perform well on a variety of different datasets and can deal with high resolution images, their traditional similarity measures severely limit them in scenarios with reflective surfaces, occlusions and strong lighting changes. Recent works~\cite{tapa_mvs, pcf_mvs} try to complete these missing areas using explicit plane priors. However, following~\cite{mvscrf}, we propose to use a CRF based regularization~\cite{bp_reloaded} of the score volume for general scenarios, which helps the system to recover correct depth measurements without relying on assumptions about the scene structure. 
\subsection{Learning-based MVS}
The concepts for supervised machine learning based MVS defined by MVSNet~\cite{mvsnet} and DeepMVS~\cite{deepmvs} were the basis for many other works which further improve upon this architecture. They utilize plane hypothesis to compute a variance metric from feature maps extracted by a CNN. The score volume is the output of a 3D convolution based neural network. RMVSNet~\cite{rmvsnet} significantly improves MVSNet~\cite{mvsnet} in terms of memory consumption by utilizing a recurrent neural network. PMVSNet~\cite{pmvsnet} learns a confidence for aggregating feature maps from the reference and source images. Recently, AttMVS~\cite{attmvsnet} utilized attention~\cite{attention_neurips} for this task. 

Other recent works adopt coarse to fine schemes to iteratively improve upon the depth prediction and to reduce memory consumption further. CVP-MVSNet~\cite{cvp_mvsnet} continuously refines the result by learning depth residuals to improve upscaled results from coarse resolution levels, similar to CasMVSNet~\cite{casmvs}. In order to optimize for efficiency in terms of memory and runtime FastMVSNet~\cite{fastmvsnet} learns to refine a sparse depth map, however this comes at the expense of the quality of the results compared to~\cite{cvp_mvsnet, casmvs}. PointMVSNet~\cite{pointmvsnet} approaches the refinement problem in 3D space by refining the point cloud in a coarse to fine manner. A related approach is SurfaceNet~\cite{surfacenet}, which operates on 3D voxel representations and is further improved by SurfaceNet+~\cite{surfacenet_plus}, which introduces a novel view selection approach that can handle sparse MVS configurations. However, methods which work in 3D space are limited in terms of resolution because of increased memory requirements. By performing a coarse to fine scheme in 2.5D,~\cite{cvp_mvsnet, casmvs} achieve a good trade-off between accuracy and computational requirements. However, there is no explicit regularization or refinement present for the end-result which reduces outliers. MVSCRF~\cite{mvscrf} employs a CRF as RNN implementation~\cite{crfasrnn}, where the CRF distribution is approximated using a simpler distribution~\cite{koltun_meanfield, crfasrnn}. While~\cite{mvscrf} is the first architecture to apply CRF regularization for learning-based MVS, we propose additional extensions to improve CRF inference for the MVS setting. 
\section{Methodology}
In the following sections we describe the implemented hierarchical network architecture incorporating the BP-layer~\cite{bp_reloaded} in detail. Furthermore, we motivate and explain our contributions, which extend the BP-layer~\cite{bp_reloaded} for use in the MVS setting.

\subsection{Model Overview} \label{sec_model_overview}
\begin{figure*}[t]
    \centering
     \includegraphics[width=1.0\columnwidth]{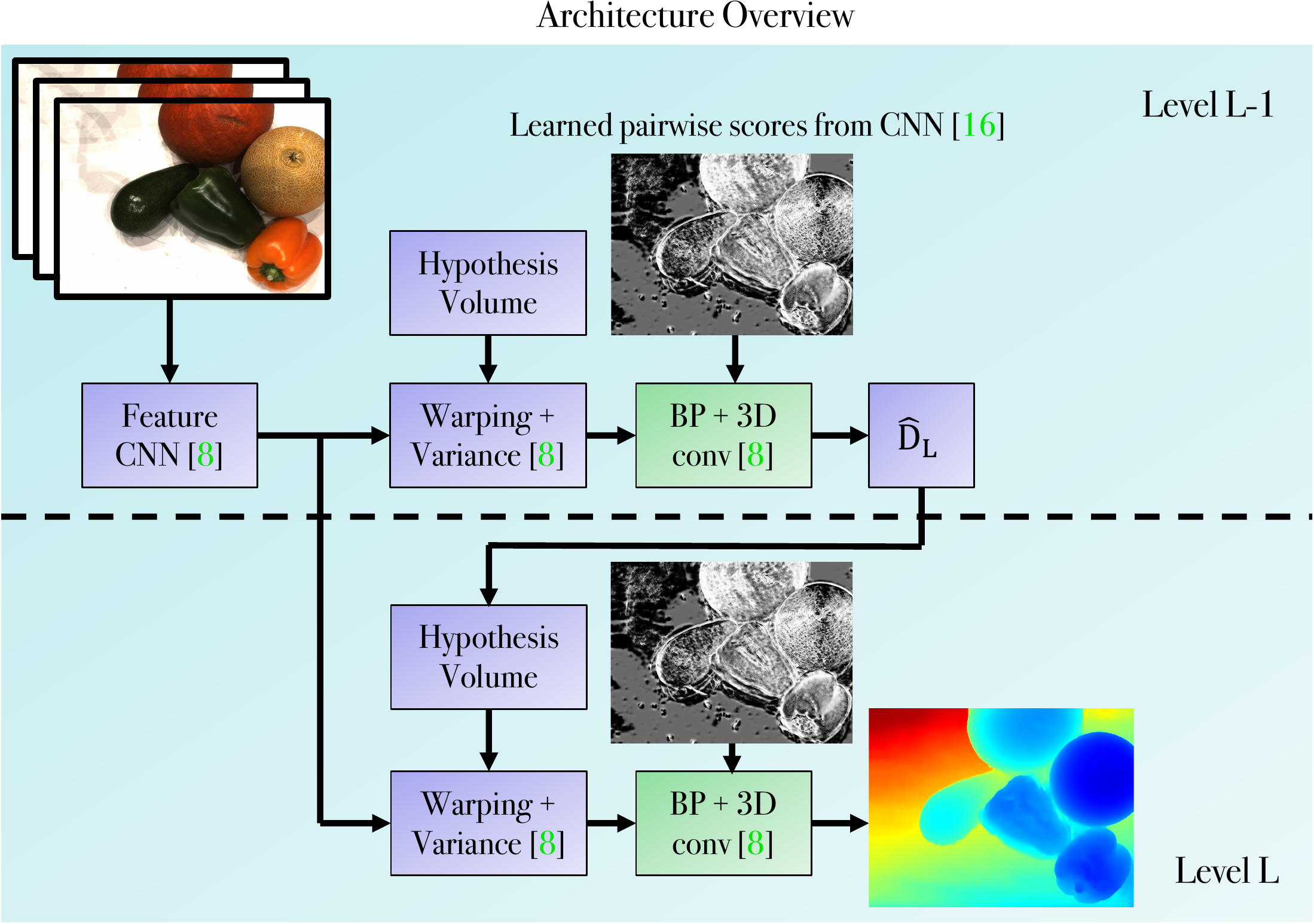}
     \includegraphics[width=1.0\columnwidth]{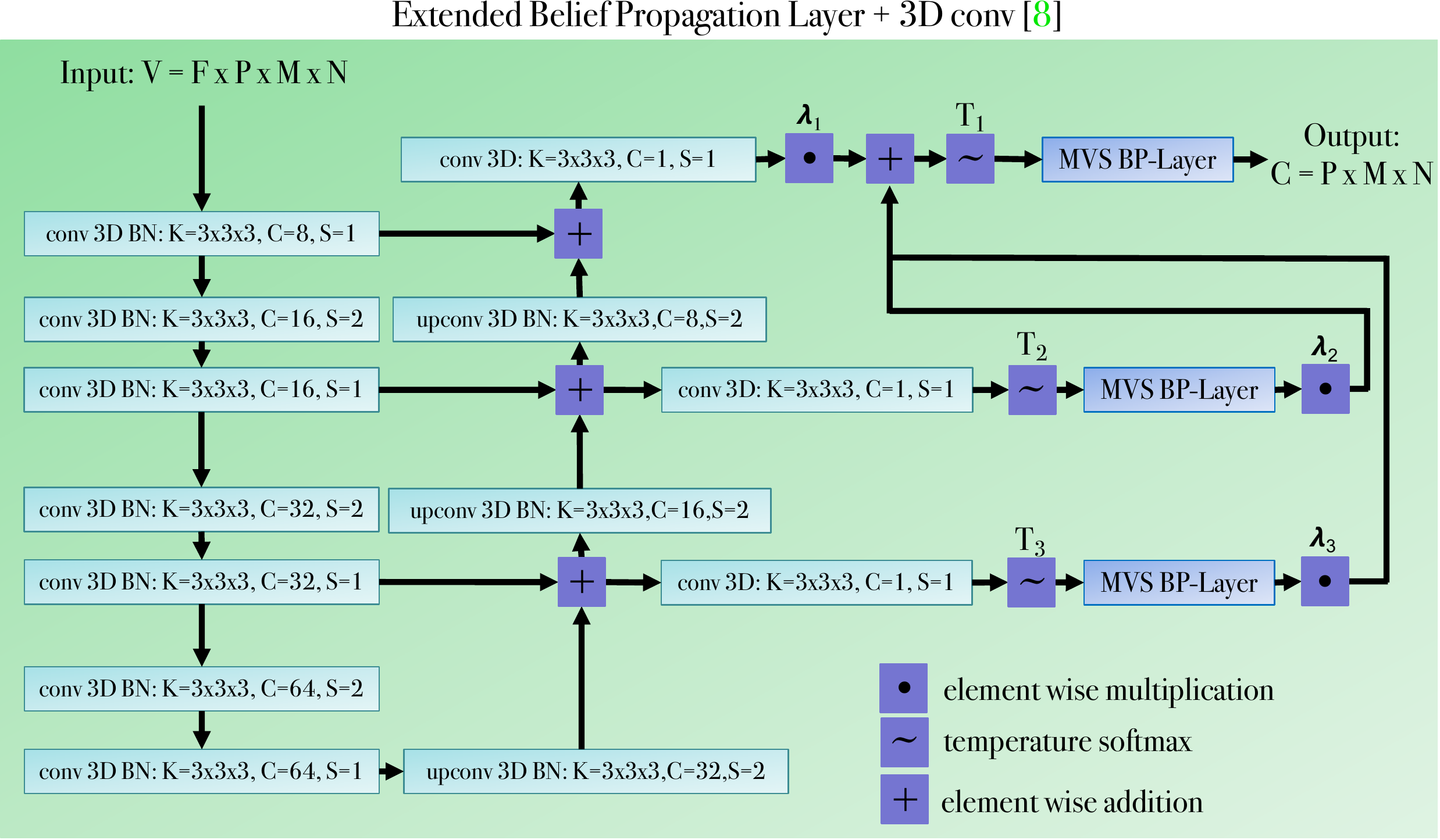}
    \caption{Left: High level overview of the model utilizing the extended BP-layer~\cite{bp_reloaded} for the MVS setting. Right: Detailed architecture of the matching network integrating the extended BP-layer. In the detailed architecture we perform trilinear interpolation to upscale the results from the lower resolution level when computing the input for the final BP-layer. In the 3D convolution description K denotes the kernel size, C the number of output channels and S the stride. }
    \label{fig_network_arch}
\end{figure*}
We work on images of size $M \times N$. Following~\cite{casmvs, mvsnet}, we discretize our search space for the depth in each pixel location into a set of labels $\mathcal L \in \{0,...,P - 1\}$, which correspond to fronto-parallel plane hypothesis. The quality of the depth estimate that each of these planes represent is quantified by a score volume $C \in \mathbb{R}^{P \times M \times N}$. One of the problems with this approach is that the computed scores in $C$ may be inconsistent in areas where occlusions, reflective surfaces and changing light conditions are present in the images, resulting in noisy or wrong depth estimates. Similar to~\cite{mvscrf}, we perform a regularization step on the score volume using a CRF. The function maximized in the CRF of~\cite{bp_reloaded} for a given label assignment $x \in \mathcal{L}^{\abs{\mathcal{V}}}$ is defined as 
\begin{equation} \label{eq_bp_energy}
    p(x) = \frac{1}{Z} \exp \Big( \sum_{i \in \mathcal{V}} g_i(x_i) + \sum_{(i,j) \in \mathcal{E}} f_{ij}(x_i, x_j)   \Big),
\end{equation}
where $\mathcal{V}$ is the set of nodes of the graphical model, $\mathcal{E}$ is the set of edges and $Z$ is a normalization constant. This function includes a unary score term $g: \mathcal{L} \to \mathbb{R}$, as well as pairwise scores $f_{ij}: \mathcal{L} \times \mathcal{L} \to \mathbb{R}$ which allow the model to penalize inconsistencies between neighbouring pixel locations. The authors of~\cite{bp_reloaded} propose a BP-layer, which performs inference in the CRF. The BP-layer is fully differentiable and can thus be integrated into any labeling based model. We provide further details on how we extended the BP-layer for the MVS setting in Section~\ref{sec_extending_bp}. 
\subsection{Network Architecture}
The first step of our MVS network is to extract features from the reference image and each of the corresponding source images. In this stage, we incorporate the multi-level feature extraction CNN of~\cite{casmvs}, which extracts $F$ feature maps. We use three resolution levels based on evaluations of~\cite{casmvs}. The output of this CNN are the feature maps $R \in \mathbb{R}^{F \times M \times N}$. The following steps are executed in the three hierarchy levels we incorporate. The source feature maps $R$ are warped according to  $P$ fronto-parallel planes~\cite{casmvs, mvsnet}, which yields a tensor $W \in \mathbb{R}^{F \times P \times M \times N}$ as the output. These warped feature maps are then used to compute the variance between the reference and all warped source feature maps. The resulting variance tensor $V \in \mathbb{R}^{F \times P \times M \times N}$ is used as the input for a matching network utilizing 3D convolutions, which outputs the final score volume $C \in \mathbb{R}^{P \times M \times N}$. Following the architecture proposed by~\cite{mvscrf}, the matching network integrates the BP-layer as a regularization component as shown in Figure~\ref{fig_network_arch}. We utilize three BP-layers on different scales in the matching network and apply a temperature softmax~\cite{bengio_dl_book} to the unary inputs as follows 
\begin{equation}\label{eq_temp_softmax}
\sigma (x(p), T) = \frac{\exp{\left(\frac{x(p)}{T}\right)}}{\sum_{j=0}^{P-1} \exp{\left(\frac{x(j)}{T}\right)}},~ 0 \leq p < P,
\end{equation}
using learned parameters $T_1$, $T_2$, $T_3$ for the respective levels. The input to the BP-layer on the highest scale is calculated as a weighted sum with weights $\lambda_1$, $\lambda_2$ and $\lambda_3$ which are learned. Further, we train three different pixel-wise pairwise networks using the UNet architecture described in~\cite{bp_reloaded} for each of the BP layers incorporated into the matching network. We train separate matching networks for each of the hierarchy levels as it has been shown by~\cite{casmvs} that this improves performance.  The resulting depth map $\hat D \in \mathbb{R}^{M \times N}$ is then computed as 
\begin{equation} \label{eq_depth_comp}
\begin{split}
    \hat D(i,j) = \sum_{p=0}^{P-1} \Tilde{C}(i,j,p)  H(i,j,p), \\ 0 \leq i < M,~0 \leq j < N,
\end{split}
\end{equation} 
where $\Tilde{C} = \sigma(C, 1)$ is the softmax normalized score volume and $H \in \mathbb{R}^{P \times M \times N}$ is a volume containing the depth hypothesis for each pixel. Following the hierarchical architecture of~\cite{casmvs}, we compute the hypothesis for the next level using the result from the current level as described in Section~\ref{sec_quant_label_jumps}. This means that the hypothesis volume contains the same labels for each pixel in the first hierarchy level and contains different labels per pixel in further levels. 
\subsection{Extension of the BP-layer} \label{sec_extending_bp}
We now provide a detailed description of our extensions to the BP-layer~\cite{bp_reloaded}, which normalize CRF label jumps for the MVS setting and allow for factional jumps in the pairwise score computation. Further, we describe how we automatically compute the depth hypothesis discretization beyond the initial stage. We show the performance improvements of our contributions in Section~\ref{sec_experiments}.
\subsection{Label jump normalization} \label{sec_quant_label_jumps}
We use the differentiable BP-layer proposed in~\cite{bp_reloaded} employing the max-sum variant of belief propagation. The advantages of using belief propagation as the inference algorithm for the CRF used in the matching network are that it is able to model long range interactions, can be efficiently implemented and is interpretable~\cite{bp_reloaded}. 
In our case the unary scores $g \in \mathbb{R}^{P \times M \times N}$ defined in Equation~\eqref{eq_bp_energy} are inputs to the BP-layer from the matching CNN after softmax activation as shown in Figure~\ref{fig_network_arch}. The term $f_{i,j}(s, t)$ introduced in Equation~\eqref{eq_bp_energy} represents the pairwise scores from label $s$ to label $t$. We apply the BP-layer as proposed in~\cite{bp_reloaded}. In the standard two-view stereo case the labels $s$ and $t$ represent disparities corresponding to horizontal displacements measured in pixels independent of scene scale. In the plane sweeping MVS case, each source and target label $s,t$ in the score volume represents the depth hypothesis $d_s, d_t \in \mathbb{R}$ of a different fronto-parallel plane which is dependent on the scene scale. If we now want to learn the pairwise score term in the MVS setting, we need to normalize depth jumps first, such that it can be applied for any scale. To this end, we apply a normalization $\mathcal{N}(d): \mathbb{R} \to \mathbb{R}$ to $d_s$ and $d_t$. We define $\mathcal{N}(d)$, based on the expected 3D error~\cite{molton_error_prop, pcf_mvs, ltvre} in the depth dimension. For a given depth value $d$, this error measure is defined as
\begin{equation} \label{eq_expected_3d_error}
\mathcal{N}(d) = \sigma_p  \frac{d^2}{\Tilde F  \Tilde{b}}  \sqrt{2}
\end{equation}
where $\Tilde{b} = \sum_{i=1}^S b_i$ represents the average baselines from all used $S$ source images, $\Tilde F$ is the focal length of the reference camera and $\sigma_p$ is a pixelwise uncertainty which we set to $1$ in all experiments. We then normalize our source and target depths using this error measure by
\begin{equation} 
\begin{split}
    \Tilde{d_s} = \frac{d_s}{\mathcal{N}(d_s)} =  \frac{d_s}{d_{s}^{2}}  \frac{\Tilde F  \Tilde{b}}{\sigma_p \sqrt{2}} = \frac{1}{d_s} \frac{\Tilde F  \Tilde{b}}{\sigma_p \sqrt{2}},
\end{split}
\end{equation}
where the tilde denotes normalized depth values. The normalized difference from one depth to another is then computed as 
\begin{equation} \label{eq_label_jump}
\begin{split}
    \Tilde{d_s} - \Tilde{d_t} = \Big(\frac{1}{d_s}  - \frac{1}{d_t} \Big)  \frac{\Tilde F  \Tilde{b}}{\sigma_p \sqrt{2}}.
\end{split}
\end{equation}
Using Equation~\eqref{eq_label_jump}, we observe that we first invert both depth hypothesis and calculate the difference in the inverse depth. Afterwards, we scale the difference by a factor of $\frac{\Tilde F \Tilde{b}}{\sqrt{2}}$ to account for the differing scene scales in the MVS setting. This is also related to how a disparity in the two-view setting is calculated from depth. Thus, we are actually operating on a scaled inverse depth. Also intuitively, calculating the distance in inverse depth makes sense as the BP-layer relies on the assumption that the gradient in the labels is constant for slanted surfaces, which is not the case when using non inverted depth maps. By incorporating the error model, this yields an extended depth distance normalization measure compared to~\cite{pcf_mvs}. Since the label jumps are now normalized using $\mathcal{N}(d)$, which is independent of the scene scale~\cite{pcf_mvs}, the pairwise score function is also scale independent and can be applied in the same manner for any scene. Note that $\Tilde{d_s}$ and $\Tilde{d_t}$ are real values and they are thus not explicit in our label set $\mathcal{L}$. We tackle this problem by performing linear interpolation when computing the pairwise score. 
\begin{figure}[t]
    \centering
     \includegraphics[width=1.0\columnwidth]{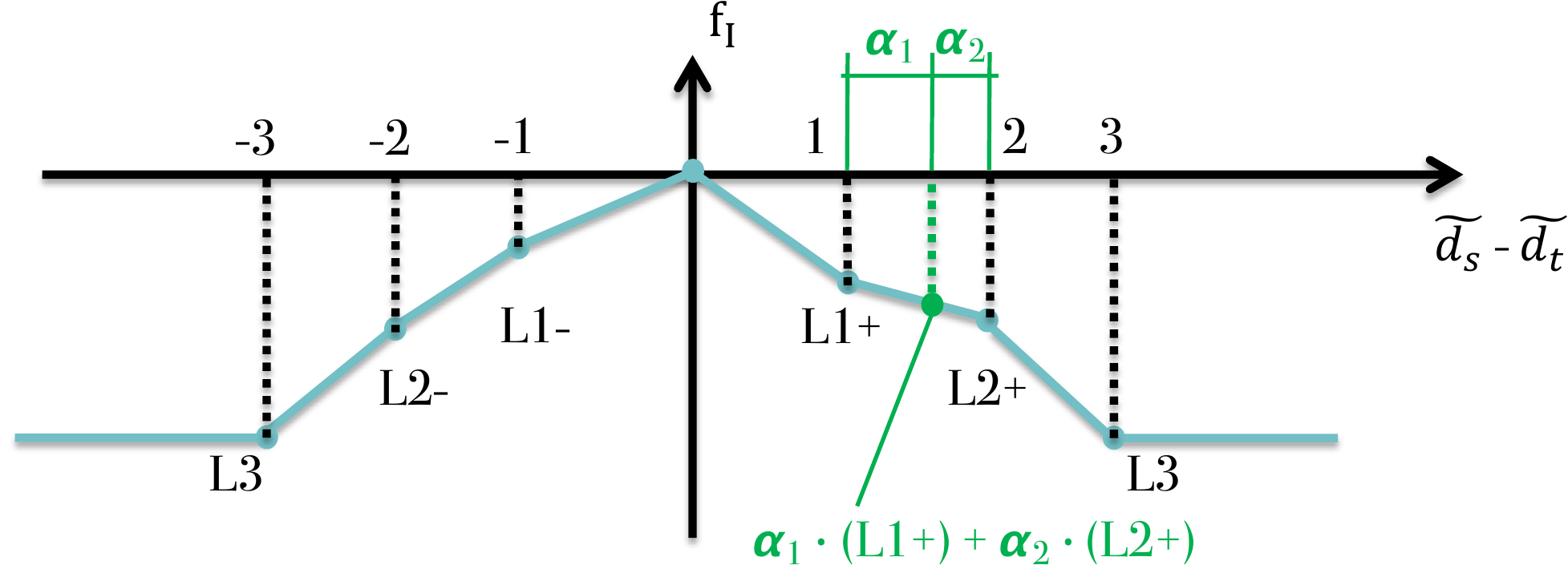}
    \caption{Computation of the pairwise score $f_I$ from learned parameters $\pm L1, \pm L2, L3$ with linear interpolation.}
    \label{fig_pairwise_score}
\end{figure}
\subsection{Pairwise score interpolation} \label{sec_pairwise_interpol}
When learning the pairwise score term we estimate the 5 parameters $\pm L_1, \pm L_2, L_3$ similar as~\cite{bp_reloaded}. These model positive and negative label jumps of quantity $1, 2$ and $\geq 3$ respectively. $L_1$ and $L_2$ can be different for positive or negative jumps while $L_3$ has the same value for both. Hence, we define the pairwise score function for two labels $s$ and $t$ by:
\begin{equation}
    f(s, t) = \Hat{f}(s - t) = \begin{cases}
        L_i~\text{for: } \abs{s - t} = i \\
        L_3~\text{for: } \abs{s - t} \geq 3.
    \end{cases}
\end{equation}
The way $\Tilde{d_s} - \Tilde{d_t}$ is defined in Equation~\eqref{eq_label_jump} also implies that there are fractional jumps for our normalized depths. Consequently, we perform a linear interpolation between our learned discrete parameters. Hence, we define our interpolated pairwise score function as
\begin{equation} \label{eq_pairwise_interpolate}
\begin{split}
   f_I(\Tilde{d_s} - \Tilde{d_t}) = \alpha_1 \Hat{f}(\floor{\Tilde{d_s} - \Tilde{d_t}}) + \alpha_2 \Hat{f}(\ceil{\Tilde{d_s} - \Tilde{d_t}}) \\ \alpha_1 + \alpha_2 = 1,~\alpha_1,\alpha_2 \geq 0
\end{split}
\end{equation}
This step can be integrated into the backward path of the BP-layer~\cite{bp_reloaded} by following the chain rule and applying 
\begin{equation} \label{eq_pairwise_backward}
\begin{split}
    \dv{\Hat{f}}(\floor{\Tilde{d_s} - \Tilde{d_t}}) \pluseq \alpha_1  z \\
    \dv{\Hat{f}}(\ceil{\Tilde{d_s} - \Tilde{d_t}}) \pluseq \alpha_2  z
\end{split}
\end{equation} 
where $z$ is the incoming gradient as defined in~\cite{bp_reloaded}. Intuitively, we distribute the incoming gradient based on the weights $\alpha_1$ and $\alpha_2$ as shown in Equation~\eqref{eq_pairwise_backward} using the notation of~\cite{bp_reloaded}. Performing this interpolation allows us to more accurately represent the pairwise score function for our learning setting, where fractional normalized depth jumps are a common occurrence. We also visualize the pairwise score computation in Figure~\ref{fig_pairwise_score}. 
\subsection{Automatic calculation of depth hypothesis sampling interval} \label{sec_auto_interval}
We compute the hypothesis volume for the initial hierarchy level as proposed in~\cite{casmvs}. However, the error measure described in Section~\ref{sec_quant_label_jumps} allows us to compute the label tensor from the depth estimate after the initial hierarchy level automatically, as opposed to manually defining a scale factor which is done in~\cite{casmvs}. We use $\mathcal{N}(d)$ to normalize our depth hypothesis, thus we want the resolution of the hypothesis volume $H$ to be at least as fine grained as this normalization factor, such that our pairwise score function can capture the corresponding depth jumps. Hence, from the upscaled depth estimate $\hat D \in \mathbb{R}^{M \times N}$ of the initial level we compute our pixelwise intervals $I \in \mathbb{R}^{M \times N}$ as
\begin{equation}
    I(i,j) = \frac{\mathcal{N}(\hat D(i,j))}{2},~ 0 \leq i < M,~0 \leq j < N.
\end{equation}
When considering the number of depth hypothesis $P$ for a given hierarchy level we then compute the label tensor $H \in \mathbb{R}^{P \times M \times N}$ for every element as
\begin{equation} \label{eq_auto_interval}
\begin{split}
 H(i,j, p) = \hat D(i,j) - \Big(\frac{P}{2} - p\Big) I(i,j),  \\
 0 \leq i < M,~0 \leq j < N,~0 \leq p < P.
\end{split}
\end{equation}
\section{Training} \label{sec_training}
We use the Adam~\cite{adam_optimizer} optimizer to train the PyTorch~\cite{pytorch} implementation of the network with the Huber loss~\cite{huber_loss} function 
\begin{equation} \label{eq_huber_loss}
    \mathcal{H}(\hat d, d^\ast) = \begin{cases}
    \frac{1}{2} (\hat d - d^\ast)^2 & \text{for}~\abs{\hat d - d^\ast} \leq \epsilon \\
    \epsilon\abs{\hat d - d^\ast} - \frac{\epsilon^2}{2} & \text{else}
    \end{cases}
\end{equation} on the resulting pixelwise depth predictions $\hat d$ and the ground truth depth $d^\ast$. We use $\epsilon=1$ in all of our experiments. The loss in each hierarchy level is calculated as
\begin{equation}
L_h = \frac{1}{MN}  \sum_{i,j=0}^{M-1, N-1} \mathcal{H}(\hat D(i,j), D^\ast(i,j))
\end{equation} 
where $h \in \{0,1,2\}$ is the hierarchy level. $\hat D \in \mathbb{R}^{M \times N}$ and $D^\ast \in \mathbb{R}^{M \times N}$ represent the estimated and ground truth depth map for the respective level. Following the work of~\cite{casmvs}, the final network loss is calculated as a weighted sum $L = \sum_{h=1}^3 \alpha_i  L_h$. We train our network with a batch size of $1$. For our ablation study on the extensions of the BP-layer~\cite{bp_reloaded}, we trained the network on a subset of the DTU training set of~\cite{mvsnet} for 7 epochs using a learning rate of $10^{-4}$. We evaluate on the full validation set from~\cite{mvsnet} after every epoch and use the epoch with the lowest error on the \SI{2}{\milli \metre} threshold for our results in Table~\ref{tab_ablation}. For the evaluations of depth maps and point clouds we trained on the full DTU training set by adding the validation set of~\cite{mvsnet} with a learning rate of $10^{-3}$ for 10 epochs and then continue to train for 4 epochs with a learning rate of $10^{-4}$. For our experiments on Tanks and Temples~\cite{tanksandtemples} and ETH3D~\cite{eth3d}, we fine-tune the model trained on the full DTU training set using the BlendedMVS~\cite{blended_mvs} training set. We train for additional 7 epochs using a learning rate of $10^{-4}$. During training we use $2$ source images in addition to the reference, while we use $4$ source images during inference. Further, we provide the used image resolution $M \times N$, number of hypothesis per level $\# H$, memory consumption, runtime and fusion paramters for training and inference in Table~\ref{tab_param_settings}. For the fusion parameters we first state the number of views $n$ which have to satisfy that the forward-backward reprojection error is less than $\tau$~\cite{mvsnet}. We use the camera parameters provided by~\cite{mvsnet} in our experiments.
\section{Experiments}\label{sec_experiments}
In the following sections we describe our evaluation procedures on the DTU~\cite{dtu}, Tanks and Temples~\cite{tanksandtemples} and ETH3D-low resolution~\cite{eth3d} datasets. Furthermore, we provide an ablation study to quantify the improvements gained by integrating the extended BP-layer~\cite{bp_reloaded}. Additionally, we evaluate the resulting depth maps on DTU~\cite{dtu} and discuss point cloud results on the datasets.
\begin{table}[t] 
\small
\centering
\setlength{\tabcolsep}{3pt}
\begin{tabular}{@{}lccccc@{}}
 \toprule 
 Dataset & $M \times N$ & \#H & Mem. & t &  $n,\tau$  \\ \midrule
 DTU~\cite{dtu} train & $640 \times 512$ & 96,32,8 & 8.9 & 1.3 & -, -  \\ 
 DTU~\cite{dtu} test & $1152 \times 864$ & 128,32,8 & 6.6 & 2.0 & 3, 0.25   \\
 Blended~\cite{blended_mvs} train & $768 \times 576$ & 96,32,8 & 10.9 & 1.6 & -, -   \\
 T \& T~\cite{tanksandtemples} inter. & $1920 \times 1056$ & 96,32,8 & 10.9 & 2.7 & 5, 0.50  \\
 T \& T~\cite{tanksandtemples} adv. & $1920 \times 1056$ & 96,32,8 & 10.9 & 2.7 & 3, 0.25  \\
 ETH3D~\cite{eth3d} test & $928 \times 512$ & 128,32,8 & 3.6 & 1.0 & 3, 0.10  \\ 
 \bottomrule
 \end{tabular}
\caption{The parameter settings for our experiments, where column \#H represents the number of depth hypothesis for the three hierarchy levels and $n,\tau$ are the fusion parameters described in Section~\ref{sec_pcl_fusion}. We also provide the used memory in GB and runtime $t$ in seconds.}
\label{tab_param_settings}
\end{table}
\begin{table}[t] 
\small
\centering
\setlength{\tabcolsep}{3pt}
\begin{tabular}{@{}lccc|cccc@{}}
 \toprule
 Method & norm. & $f_I$ & $H_A$ & \SI{2}{\milli \metre} & \SI{4}{\milli \metre} & \SI{8}{\milli \metre} & \SI{20}{\milli \metre} \\ \midrule
BP-MVSNet & \checkmark & -  & - & 24.20 & 14.34 & 9.92 & 6.53  \\ 
BP-MVSNet & \checkmark & \checkmark & - & 23.77 & \textbf{13.53} & \textbf{9.02} & \textbf{5.85}  \\  
BP-MVSNet & \checkmark & \checkmark & \checkmark & \textbf{23.21} & 14.04 & 9.38 & 5.95  \\  \hline
 CasMVSNet~\cite{casmvs} & - & - & - & 24.51 & 16.18 & 11.93 & 8.41   \\  
 \bottomrule
 \end{tabular}
\caption{Ablation study on the DTU validation set depth maps of~\cite{mvsnet}. All methods have been trained on a representative subset of the full training-set. We report the average percentage of pixels where the error is larger than a given threshold, for thresholds in the range of \SI{2}{\milli \metre} -  \SI{20}{\milli \metre} (lower is better). Column norm. indicates that we enable normalized label jumps and column $f_I$ indicates that we use interpolated pairwise scores. With column $H_A$ we indicate that we enable the automatic computation of the label discretizations after the initial stage.}
\label{tab_ablation}
\end{table}
\subsection{DTU dataset}
The 128 scenes from the DTU dataset~\cite{dtu} capture objects placed on a table using a full image resolution of $1600 \times 1200$. The ground truth is provided by a structured light scanner. The evaluation metrics for DTU are a completeness metric, which is the average distance from ground truth points to the nearest reconstructed point and an accuracy metric which is the average distance from the reconstructed points to the nearest ground truth point. Furthermore, large outliers and points not in the observability mask are filtered~\cite{dtu}. 
\begin{table}[t]
\small
\centering
\begin{tabular}{@{}lccccc@{}}
 \toprule
 Method & \SI{2}{\milli \metre} & \SI{4}{\milli \metre} & \SI{8}{\milli \metre} &  \SI{20}{\milli \metre} \\ \midrule
 CVP-MVSNet~\cite{cvp_mvsnet} & 28.99 & 21.78 & 16.16 & 9.48 \\  
 CasMVSNet~\cite{casmvs} & 24.84 & 19.74 & 16.14 & 11.56  \\  
 BP-MVSNet & \textbf{21.79} & \textbf{15.88} & \textbf{11.57} & \textbf{7.70}  \\  
 \bottomrule
 \end{tabular}
\caption{Comparison of DTU~\cite{dtu} depth maps on the test set of~\cite{mvsnet}. We list the average percentage of pixels with an error larger than \SI{2}{\milli \metre} -  \SI{20}{\milli \metre} (lower is better).}
\label{tab_depth_map_comp}
\vspace{-1em}
\end{table} 
\begin{table}[t] 
\centering
\begin{tabular}{@{}lccc@{}}
 \toprule
 Method & overall & acc. & comp. \\ \midrule
 MVSNet~\cite{mvsnet} & 0.462 & 0.396 & 0.527   \\  
 R-MVSNet~\cite{rmvsnet} & 0.422 & 0.385 & 0.459   \\  
 MVSCRF~\cite{mvscrf} & 0.398 & 0.371 & 0.426   \\  
 P-MVSNet~\cite{pmvsnet} & 0.420 & 0.406 & 0.434   \\  
 Fast-MVSNet~\cite{fastmvsnet} & 0.370 & 0.336 & 0.403   \\  
 CasMVSNet~\cite{casmvs} & 0.355 & 0.325 & 0.385   \\  
 Att-MVS~\cite{attmvsnet} & 0.356 & 0.383 & 0.329   \\  
 CVP-MVSNet~\cite{cvp_mvsnet} & 0.351 & \textbf{0.296} & 0.406  \\  \hline
 BP-MVSNet (ours) & \textbf{0.327} & 0.333 & \textbf{0.320}  \\ 
 \bottomrule
 \end{tabular}
\caption{Overall, completness and accuracy (lower is better) results on the point clouds of the DTU~\cite{dtu} test set of~\cite{mvsnet}. }
\label{tab_dtu_point_cloud}
\vspace{-1em}
\end{table}
\begin{figure*}[t]
    \centering
     \includegraphics[width=2.0\columnwidth]{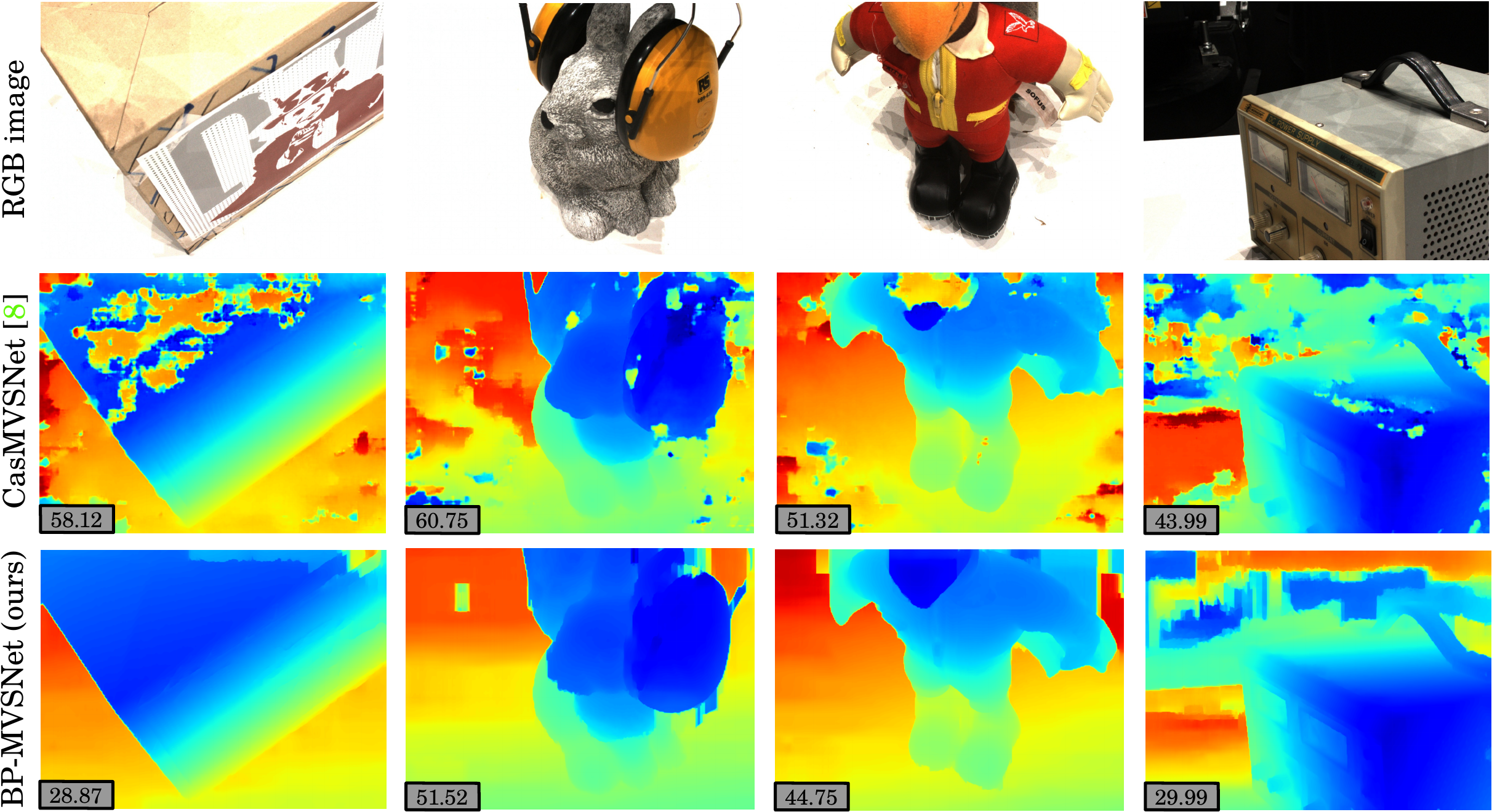}
    \caption{RGB image and depth map results from BP-MVSNet and CasMVSNet~\cite{casmvs}. The depth maps include the percentage of pixels with an error larger than \SI{2}{\milli \metre} compared to the ground truth (lower is better).}
    \label{fig_depth_maps_full_dtu}
\end{figure*}
\begin{table*}[t] 
\centering
\scriptsize
\setlength{\tabcolsep}{4pt}
\begin{tabular}{@{}lcccccccccccccccc@{}}
 \toprule
 & \multicolumn{9}{c}{intermediate} & \multicolumn{7}{c}{advanced} \\ 
 \cmidrule{2-17} 
 Method & F & Fam. & Franc. & Horse & Light. & M60 & Panther & Playg. & Train & F & Audit. & Ballr. & Courtr. & Museum & Palace & Temple \\ \midrule
 COLMAP~\cite{colmap_mvs} & 42.14 & 50.41 & 22.25 & 25.63 & 56.43 & 44.83 & 46.97 & 48.53 & 42.04 & 27.24 & 16.02 & 25.23 & 34.70 & 41.51 & 18.05 & 27.94 \\ 
 ACMM~\cite{acmm} & 57.27 & 69.24 & 51.45 & 46.97 & 63.20 & 55.07 & 57.64 & 60.08 & 54.48 & 34.02 & 23.41 & 32.91 & \textbf{41.17} & 48.13 & 23.87 & 34.60  \\ 
 PCF-MVS~\cite{pcf_mvs} & 55.88 & 70.99 & 49.60 & 40.34 & 63.44 & \textbf{57.79} & 58.91 & 56.59 & 49.40 & \textbf{35.69} & \textbf{28.33} & \textbf{38.64} & 35.95 & 48.36 & 26.17 & \textbf{36.69}  \\ 
 MVSNet~\cite{mvsnet} & 43.48 & 55.99 & 28.55 & 25.07 & 50.79 & 53.96 & 50.86 & 47.90 & 34.69 & - & - & - & - & - & - & -  \\  
 R-MVSNet~\cite{rmvsnet} & 48.40 & 69.96 & 46.65 & 32.59 & 42.95 & 51.88 & 48.80 & 52.00 & 42.38 & 24.91 & 12.55 & 29.09 & 25.06 & 38.68 & 19.14 & 24.96  \\  
 MVSCRF~\cite{mvscrf} & 45.73 & 59.83 & 30.60 & 29.93 & 51.15 & 50.61 & 51.45 & 52.60 & 39.68 & - & - & - & - & - & - & -  \\  
 P-MVSNet~\cite{pmvsnet} & 55.62 & 70.04 & 44.64 & 40.22 & \textbf{65.20} & 55.08 & 55.17 & 60.37 & 54.29 & - & - & - & - & - & - & -   \\  
 Fast-MVSNet~\cite{fastmvsnet} & 47.39 & 65.18 & 39.59 & 34.98 & 47.81 & 49.16 & 46.20 & 53.27 & 42.91 & - & - & - & - & - & - & -  \\  
 Att-MVS~\cite{attmvsnet} & \textbf{60.05} & 73.90 & \textbf{62.58} & 44.08 & 64.88 & 56.08 & \textbf{59.39} & \textbf{63.42} & \textbf{56.06} & 31.93 & 15.96 & 27.71 & 37.99 & \textbf{52.01} & \textbf{29.07} & 28.84  \\  
 CVP-MVSNet~\cite{cvp_mvsnet} & 54.03 & 76.50 & 47.74 & 36.34 & 55.12 & 57.28 & 54.28 & 57.43 & 47.54 & - & - & - & - & - & - & -  \\ \hline
 CasMVSNet~\cite{casmvs} & 56.84 & 76.37 & 58.45 & 46.26 & 55.81 & \underline{56.11} & \underline{54.06} & 58.18 & 49.51 & 31.12 & 19.81 & \underline{38.46} & 29.10 & \underline{43.87} & 27.36 & 28.11  \\  
 BP-MVSNet(ours) & \underline{57.60} & \textbf{77.31} & \underline{60.90} & \textbf{47.89} & \underline{58.26} & 56.00 & 51.54 & \underline{58.47} & \underline{50.41} & \underline{31.35} & \underline{20.44} & 35.87 & \underline{29.63} & 43.33 & \underline{27.93} & \underline{30.91} \\ 
 \bottomrule
 \end{tabular}
\caption{F-score (higher is better) results on the Tanks and Temples benchmark~\cite{tanksandtemples} intermediate and advanced test sets. The overall best results are marked as bold numbers, while the best result between BP-MVSNet and CasMVSNet~\cite{casmvs} is underlined.}
\label{tab_tanks_results}
\vspace{-1em}
\end{table*}
\begin{table*}[t] 
\centering
\small
\begin{tabular}{@{}lcccccccccccc@{}}
 \toprule
 & \multicolumn{5}{c}{test} & \multicolumn{5}{c}{train} \\ 
 \cmidrule{2-13} 
 Method & F & lake. & sandb. & stor. & stor. 2 & tunnel & F & deliv. & electro & forest & playgr. & terrains \\ \midrule
 ACMM~\cite{acmm} & 55.01 & 59.60 & \textbf{66.07} & 35.89 & 50.48 & \textbf{63.01} & 55.12 & 38.65 & \textbf{61.75} & 60.21 & \textbf{43.87} & \textbf{71.11}  \\ 
 PCF-MVS~\cite{pcf_mvs} & \textbf{57.06} & \textbf{66.85} & 62.36 & \textbf{43.32} & \textbf{52.89} & 59.86 & \textbf{57.32} & \textbf{48.61} & 56.36 & \textbf{67.24} & 43.68 & 70.70   \\  
 COLMAP~\cite{colmap_mvs} & 52.32 & 56.18 & 61.09  & 38.61 & 46.28 & 59.41 & 49.91 & 37.30 & 52.31 & 61.83 & 31.91 & 66.23   \\  
 MVSNet~\cite{mvsnet} & 38.33 & 40.49 & 57.57 & 20.52 & 33.47 & 39.63 & - & - & -  & - & - & -  \\  
 R-MVSNet~\cite{rmvsnet} & 36.87 & 42.00 & 46.99 & 24.73 & 34.83 & 35.83 & - & - & -  & - & - & -   \\  
 P-MVSNet~\cite{pmvsnet} & 44.46 & 49.27 & 49.30 & 34.35 & 39.83 & 49.54 & - & - & -  & - & - & -   \\  
 MVSCRF~\cite{mvscrf} & 28.32 & 32.16 & 44.37 & 14.66 & 21.69 & 28.73 & - & - & -  & - & - & - \\  
 Att-MVS~\cite{attmvsnet} & 45.85 & 49.36 & 51.75 & 34.83 & 43.70 & 49.63 & - & - & -  & - & - & - \\  \hline
 CasMVSNet~\cite{casmvs} & \underline{44.49} & \underline{56.38} & \underline{64.76} & 18.64 & 31.23 & 51.43 & 49.00 & 34.83 & \underline{53.10} & \underline{66.82} & 28.93 & 61.32 \\
 BP-MVSNet (ours) & 43.22 & 52.86 & 46.16 & \underline{27.25} & \underline{36.92} & \underline{52.94} & \underline{50.87} & \underline{40.10} & 49.76 & 63.03 & \underline{34.23} & \underline{67.21}  \\ 
 \bottomrule
 \end{tabular}
\caption{F-score metrics for the ETH3D low-res~\cite{eth3d} test and training set. For CasMVSNet~\cite{casmvs}, we chose the (base) results in the benchmark. The overall best results are marked as bold numbers, while the best result between BP-MVSNet and CasMVSNet~\cite{casmvs} is underlined. }
\label{tab_eth_point_cloud}
\end{table*}
\vspace{-1.2em}
\subsubsection{Ablation study} \label{sec_ablation_study}
In our ablation study, we evaluate the impact of our extensions to the BP-layer~\cite{bp_reloaded} on the performance of BP-MVSNet by comparing error metrics computed on depth maps with resolution $640 \times 512$ from the DTU validation set of~\cite{mvsnet}. The error metrics we use are the percentage of errors larger than thresholds $\{2,4,8\}$ mm between the ground truth depth map $D^\ast$ and the estimated depth map $\hat D$. We can observe in Table~\ref{tab_ablation} that adding the BP layer using the normalized depth jumps improves all of the error metrics compared to the CasMVSNet~\cite{casmvs} baseline. Enabling the interpolation step for pairwise scores further improves all of the metrics. This means that the combination of these two contributions to the BP-layer improve its performance significantly in the MVS setting. Finally, we also include the automatic computation of the label volume discretizations beyond the initial hierarchy level. This yields the lowest error on \SI{2}{\milli \metre}, thus we will use this model for all further experiments. The reason for the slight error increase with bigger thresholds is related to the dependence of the sampling interval hypothesis on the previous depth estimate. In case the previous estimate is wrong, a larger fixed interval can potentially improve results for bigger error thresholds. 
\vspace{-1.3em}
\subsubsection{Evaluation on depth maps}
\vspace{-0.3em}
The results in Table~\ref{tab_depth_map_comp} show quantitative results on depth maps of resolution $1152 \times 864$, when using our proposed model BP-MVSNet trained on DTU as described in Section~\ref{sec_training}. We compare with CasMVSNet~\cite{casmvs} and CVP-MVSNet~\cite{cvp_mvsnet} using their respective pretrained models on the same resolution with default parameter settings. This comparison allows us to measure the quality of the depth maps before performing the point cloud fusion, as even the same depth maps can yield different results in this stage depending on the fusion parameters. It can be seen that BP-MVSNet improves on all of the measured error metrics signficantly compared to the other methods. This can also be observed in Figure~\ref{fig_depth_maps_full_dtu}, where we show that the extended BP-layer is able to correctly regularize even over large areas containing inconsistent estimates, such as the top of the box in the first image. However, it can also be seen that very poor unary estimates in the matching network can also lead to artifacts in some background regions. 
\vspace{-1em}
\subsubsection{Evaluation on point clouds} \label{sec_pcl_fusion}
\begin{figure}[t]
    \centering
    \includegraphics[width=1.0\columnwidth]{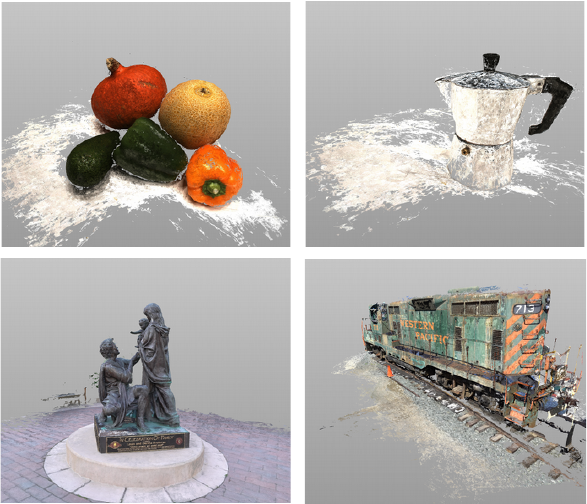}
    \caption{Top row: Point cloud results from the DTU~\cite{dtu} dataset. Bottom row: Point clouds from Tanks and Temples~\cite{tanksandtemples} scenes.}
    \label{fig_point_cloud}
    \vspace{-1.7em}
\end{figure}

\vspace{-0.5em}
For the point cloud fusion step performed for evaluations on point clouds, we utilize the method of~\cite{mvsnet}. The fusion parameters are set according to Table~\ref{tab_param_settings}, where $\tau$ denotes the forward-backward reprojection error threshold and $n$ are the number of images which have to be consistent with respect to $\tau$. Additionally we set the maximum relative depth difference parameter for the fusion of~\cite{mvsnet} to $0.01$. In Table~\ref{tab_dtu_point_cloud}, we provide the results of the DTU point cloud evaluation~\cite{dtu} on the test set of~\cite{mvsnet}. We set the point cloud fusion parameters as described in Table~\ref{tab_param_settings}. We compare our results with other state-of-the-art works for learning based MVS. It can be seen that BP-MVSNet outperforms all other methods both in terms of accuracy as well as completeness. In the point cloud results visualized in Figure~\ref{fig_point_cloud}, one can see that the extended BP-layer~\cite{bp_reloaded} was able to regularize the depth maps such that we get complete results, even on untextured and reflective surfaces, as seen on the coffee can.
\vspace{-1.7em}
\subsection{Tanks and Temples dataset}
The Tanks and Temples~\cite{tanksandtemples} dataset consists of $14$ real-world outdoor and indoor scenes, capturing objects, buildings and rooms, where the ground truth has been acquired by a laser scanner. The full image resolution is $1920 \times 1056$. This dataset includes many challenging scenes where reflections and occlusions are present. In this section, we compare the results of BP-MVSNet on the Tanks and Temples~\cite{tanksandtemples} benchmark with other state of the art methods. The metrics used in the Tanks and Temples~\cite{tanksandtemples} dataset are a precision score $Pr$, which is the percentage of reconstructed points with a distance $<T$ to the nearest ground truth point. Further, the recall $Re$ gives the percentage of ground truth points where the closest distance to a reconstructed point is $<T$. The F-score~\cite{tanksandtemples} is then computed as $F = \frac{2 Pr Re}{Pr + Re}$.
\subsubsection{Benchmark results}
 In Table~\ref{tab_tanks_results}, we present the results of BP-MVSNet on the Tanks and Temples benchmark~\cite{tanksandtemples}. We also provide the fusion parameters in Table~\ref{tab_param_settings}. For the Horse dataset of the intermediate scenes we increased $\tau$ to 1.0 as the base of the statue contains many reflections. It can be observed that BP-MVSNet achieves competitive results, outperforming the base architecture~\cite{casmvs} in terms of the mean F-score metric on the advanced and intermediate sets. In Figure~\ref{fig_point_cloud}, we visualize the resulting point clouds. It can be observed that even difficult scenes containing surfaces that reflect the sky such as the base of the statue are quite complete. Furthermore, as seen in Figure~\ref{fig_point_cloud}, smaller details such as the rails on the train are preserved.
\subsection{ETH 3D dataset}
The ETH3D low resolution~\cite{eth3d} dataset consists of $10$ outdoor and indoor scenes of varying locations from a forest to a storage room. The full image resolution for this dataset is $928 \times 512$. Similar to Tanks and Temples~\cite{tanksandtemples}, the fused point clouds are evaluated based on a laser-scanner ground truth, in terms of accuracy, completeness and F-score.
\vspace{-1em}
\subsubsection{Benchmark results}
\vspace{-0.5em}
In Table~\ref{tab_eth_point_cloud}, we provide results for the ETH3D-low-resolution many view~\cite{eth3d} benchmark. It can be seen that we achieve competitive results among learning-based methods such as MVSNet~\cite{mvsnet} and CasMVSNet~\cite{casmvs}. Further, we did not train our network on the training set images of this dataset as described in Section~\ref{sec_training}. However, we can also observe that all of the learning based methods are outperformed by the traditional methods PCF-MVS~\cite{pcf_mvs} and ACMM~\cite{acmm} on this dataset. Compared to the base architecture CasMVSNet~\cite{casmvs}, we achieve better scores on some datasets such as storage room and terrains, while the score for other datasets such as sandbox or electro is lower, which results in a slightly worse result on the test and a slightly better one on the training set.
\vspace{-1em}
\section{Conclusion}
\vspace{-0.5em}
In this work, we have proposed BP-MVSNet, a CNN based MVS system, employing a CRF regularization layer based on belief propagation~\cite{bp_reloaded}. In order to optimize the performance of the BP-layer~\cite{bp_reloaded} for the MVS setting, we have made three core contributions: (i) Utilizing a scale agnostic term for normalizing label jumps and (ii) implementing a differentiable interpolation step in the pairwise score computation. (iii) Further, we automatically choose the discretization in our hypothesis volume after the initial stage. These contributions improve the performance of the baseline architecture~\cite{casmvs}, as seen in our quantitative results presented in Section~\ref{sec_experiments}, where we achieve state-of-the-art results on both the DTU~\cite{dtu} and Tanks and Temples~\cite{tanksandtemples} datasets. Future work could involve the inclusion of additional information, such as the surface normals as additional guidance for the BP-layer. \textbf{Acknowledgement:} This work was supported by Pro$^2$Future (FFG, Contract No. 854184).

{\small
\bibliographystyle{ieee}
\bibliography{refs}
}

\end{document}